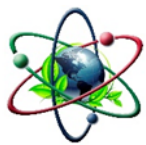

# Middle East Journal of Science

https://dergipark.org.tr/mejs

**MEJS**

# OPTIMIZED RELAY LENS DESIGN FOR HIGH-RESOLUTION IMAGE TRANSMISSION IN MILITARY TARGET DETECTION SYSTEMS

*Burak ÇELİK*[\*,1], *Kıvanç DOĞAN*[2], *Ezgi TAŞKIN*[3], *Ayhan AKBAL*[4], *Ahmet ORHAN*[5]

[1] Kocaeli University, Electronic and Communication Enginnering Orcid[1]: https://orcid.org/0000-0002-3204-5444
[2] Fırat University, Electrical and Electronics Enginnering, Orcid[2]: https://orcid.org/0000-0003-4832-1412
[3] Kocaeli University, Electrical Enginnering Orcid[3]: https://orcid.org/0000-0002-7230-3401
[4] Fırat University, Electrical and Electronics Enginnering, Orcid[4]: https://orcid.org/0000-0001-5385-9781
[5] Fırat University, Electrical and Electronics Enginnering, Orcid[5]: https://orcid.org/0000-0003-1994-4661
\* Corresponding author; burak.celik@kocaeli.edu.tr

**Abstract:** *The design and performance analysis of relay lenses that provide high-performance image transmission for target acquisition and tracking in military optical systems. Relay lenses are critical components for clear and lossless image transmission over long distances. In this study, the optical performance of a relay lens system designed and optimized using ZEMAX software is investigated in detail. The analysis focuses on important optical properties such as modulation transfer function (MTF), spot diagrams, Seidel diagram, field curvature and distortion. The results show that the lens has significant potential in military applications for target detection and tracking with high resolution and low aberration.*

## 1. Introduction

Military optical systems provide high-performance and reliable monitoring for target identification and tracking in critical missions. These systems have become indispensable in modern warfare, where the ability to process and analyze real-time visual data can determine the success or failure of operations. By combining advanced optical technologies with robust design methodologies, military systems aim to deliver precise and effective solutions for a variety of applications. In this context, optical components must ensure exceptional image clarity, resolution, and durability to withstand challenging operational environments. Electro-optical systems, which have critical features such as simultaneous image transmission, stand out in military fields by allowing intelligence, reconnaissance, surveillance, and targeting. These systems empower military personnel to detect, identify, and track targets across long distances, often in complex and dynamic scenarios. Moreover, many modern armies around the world are investing in Augmented Reality (AR) and Virtual Reality (VR) tools to elevate their systems, gain superiority over opposing forces, and prevent losses on the battlefield [1]. In these systems, relay lenses are essential components for transmitting images clearly and without loss over long distances. These lenses ensure the integrity and fidelity of transmitted images, a requirement that is particularly vital in time-sensitive and mission-critical operations. The proper use of relay lenses in technologies such as thermal cameras is of great importance for military security [2]. Thermal imaging systems, for instance, rely heavily on relay lenses to maintain image quality, enabling operators to detect threats even in low-visibility conditions. In systems like laser rangefinders (LRF) and laser target designators (LTD), the use of relay lenses is critical for directing laser beams accurately toward the target and ensuring precise focusing. These systems demand optical precision to achieve accurate target engagement, particularly in environments where atmospheric conditions and distance pose significant challenges. Laser weapons can be either ground-based or space-based. Ground-based laser weapons use multiple relay mirrors in space to intercept a theater ballistic missile. The relay mirror(s) is alens or group of lenses that transmits a finite object to a distant location with a magnification of unity or bigger values. Also they are used to extend the range of high-energy laser weapons as they compensate the limiting factors due to atmospheric absorption, turbulence and the curvature of the Earth [3]. It is a lens or group of lenses that transmits a finite object to a distant location with a magnification of unity or another value [4-5]. They are used in rifle sights, military infrared imaging systems. The use of relay lenses is quite wide as rifle sights, military imaging systems, biomedical applications [6-9] and image applications [10]. This versatility highlights the importance of designing relay lenses that can meet diverse and demanding operational requirements. In this study, a relay lens was designed and optimized in ZEMAX environment for use in rifle sights. By addressing key performance parameters such as resolution, field curvature, and distortion, this work aims to contribute to the development of high-performance optical solutions tailored for military applications.

This study follows a systematic framework to examine various aspects of the research. In chapter 2, materials and methods details the methodologies and processes used in the lens design process at ZEMAX. In chapter 3, experimental study focuses on the design of the lens and their analysis in the design process. In chapter 4, design of relay lens selection deals with the refinement of the parameters and criteria for selecting suitable materials. In chapter 5, results and discussions presents the visualization of the obtained data and the related analysis. In chapter 6, recommendations provides suggestions for future work.

## 2. Material and Method

ZEMAX is a program that can assist in the modeling, analysis and design of optical systems. The interface to ZEMAX is designed to be easy to use and with a little practice can allow very fast interactive design [11]. During the design process, the determination of the optical parameters, the analysis and the evaluation of the system performance were carried out by means of this software. Hammer and Global optimization methods are used in ZEMAX software to reduce aberration. These methods are very similar and share the same functional basis [12]. During the design process, optimization was performed using the Hammer optimization algorithm in order to minimize the aberrations in the optical system.

Relay lenses are an important group of lenses used to transfer lossless and clearly from a certain distance to another optical plane. These objects usually consist of two or more lens elements. Before starting the relay lens design, basic parameters such as focal length (EFFL) and total length (TOTR) and aperture diameter were determined considering the application where the system will be used. The performance of the design was evaluated by metrics such as modulation transfer function (MTF), spot diagram and distortion analysis. Hammer optimization was applied to improve this performance. The aim was to minimize aberrations and improve image quality. Different material combinations were tried to reduce optical aberrations.

## 3. Experimental Study

This study focuses on the design and optimization of a relay lens in a ZEMAX environment specifically designed for military applications. Lens system parameters including effective focal length (EFFL), total optical path length (TOTR), aperture, and field of view (FOV) in a ZEMAX environment were determined and analyzed to meet specific system requirements.

**3.1. Optical Design and Analysis**

In this study, the relay lens designed in ZEMAX environment is optimized for military applications. During the design process, the optical parameters of the lens were carefully determined and analyzed. The requirements of the system were analyzed and parameters such as focal length (EFFL), total optical path (TOTR), aperture and field of view (FOV) were determined. It was aimed to be suitable for use on the weapon and to have a compact structure. Designed in the ZEMAX environment, the relay lens is optimized for military applications. During the design process, the optical parameters of the lens were carefully determined and analyzed. By analyzing the system requirements, parameters such as focal length (EFFL), total optical path (TOTR), aperture and field of view (FOV) were determined. It was aimed to be suitable for use on weapons and to have a compact structure. The EFFL value is approximately 9.5 mm. A short focal length provides a wider field of view. A focal length of 9.5 mm covers a wide area at close range for a compact system. A TOTR value of 20 mm provides a compact design that meets the requirements of portability and lightness. A 3 mm aperture provides sufficient light collection, but a wider aperture may be required in low light conditions. The results obtained in this process determined the focusing ability, image quality and other optical characteristics of the lens. The relay lens obtained in ZEMAX environment is designed to be used in this field. The performance of the optical system is determined by the spot diagram, field curve/distortion diagram and modulation transfer function (MTF)[13]. In this study, the analysis results of the designed lens such as MTF, PSF, Spot Diagram, Image Simulation, Seidel Diagram, Field curvature and distortion are obtained and the Shaded model is presented. MTF evaluation of image displays is often needed to evaluate objective image quality, especially in the application of image quality measurements.

**3.2. Optimization and Material Selection**

After the lens assembly of the design was completed, different materials were tested for the lenses. The materials were optimized with the help of Hammer optimization to achieve the best image quality and low aberrations. The materials that achieved high image quality, low distortion and eliminated optical aberrations were determined. These optical aberrations are astigmatism, spherical aberrations and comas. The goal of the design is to achieve the best image quality. Hammer analysis aims to find the global minimum of the design. For this reason, materials suitable for the substituted surfaces were obtained as a result of optimization by using the "Materials Catalog" during Hammer analysis.

| MATERIALS |
|---|
| LAK8 |
| LAK11 |
| KZFS6 |
| BAF51 |
| F7 |
| KZFS1 |
| N-SK2HT |
| K5G20 |
| SK14 |
| LAKL12 |

**Figure 1.** Materials obtained as a result of Hammer optimization algorithm analysis

## 4. Design of Relay Lens

In this section, we present a detailed analysis of the optical design and its performance metrics to evaluate its suitability for high-performance image transfer in military optical systems, specifically targeting detection and tracking applications. The analyses provide a comprehensive understanding of the system's imaging quality, aberration control, and overall performance across critical parameters. The evaluation begins with the Modulation Transfer Function (MTF), a crucial metric that illustrates the spatial frequency response and resolution capability of the optical system. Subsequently, the Point Spread Function (PSF) is analyzed to assess the system's energy distribution and imaging fidelity. To further complement these metrics, the Spot Diagram offers insight into the system's aberration characteristics at various field positions. The Seidel Diagram is used to dissect the contributions of primary aberrations, enabling an understanding of the root causes of optical imperfections. Moreover, the Field Curvature and Distortion analysis is undertaken to assess the system's ability to maintain a flat image plane and minimize geometric distortions. Advanced image evaluation is performed through Extended Diffraction Image Analysis, Geometric Image Analysis, and Demo Image Analysis, which collectively provide both diffraction-limited and ray-based perspectives of imaging performance. The Merit Function optimization results are included to highlight the design's adherence to performance goals, while the Optical Path Difference (OPD) Fan Analysis offers a quantitative assessment of wavefront errors across the aperture.

By combining these analyses, we provide a holistic evaluation of the designed optical system, ensuring its robustness and efficiency for demanding military applications. The results highlight the design's capability to deliver precise, high-quality imagery for critical target detection and tracking scenarios.

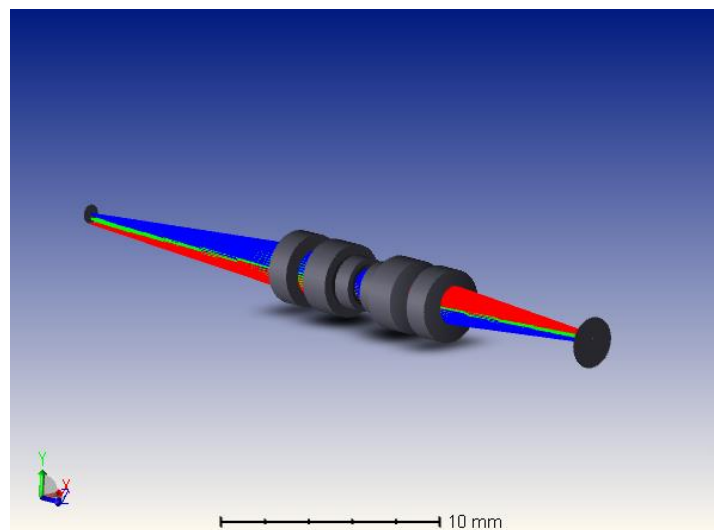

**Figure 2**. Shaded Model

As a result, Figure 2 provides a detailed visualization of the structural design and shows the arrangement and interaction of optical components within the system.
 This model shows the ray tracing simulation of the relay lens system designed in the ZEMAX environment. The shaded model in the figure presents the system's performance in focusing beams of different colors. The uniform focusing of the beams and the minimization of aberrations demonstrate the success of the optimization techniques applied during the design process.

### 4.1. Modulation Transfer Function

The analysis result in Figure 3 shows the modulation transfer function (MTF) of the lens versus spatial frequencies. Diffraction-limited resolution in the MTF plot has been a fundamental method for determining optical resolution [14]. MTF represents the transmitted contrast. The MTF plot shows how sharply the system transmits structures at spatial frequencies above the Nyquist frequency [15]. Accurate optical resolution is indispensable to ensure that instrumentation and analytical techniques give the required image quality [14]. According to the obtained MTF plot, the MTF values of the lens in the range of 0-150 cycles\mm are quite high (around 0.45). This shows that at low and medium frequencies the optical system achieves high contrast and clear images. Low frequencies usually represent large and distinct details. After 350 cycles\mm, the MTF starts to go completely to zero and at this point contrast transmission is very difficult. However, this drop in high frequencies is always to be expected.

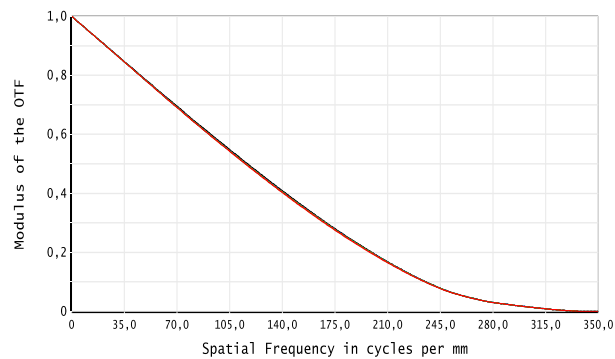

**Figure 3.** Modulation Transfer Function

As a result, Figure 3 showing the MTF plot, demonstrates the ability of the lens to effectively resolve spatial frequencies and provides a quantitative assessment of imaging performance and contrast accuracy across the field of view.

### 4.2. Point Spread Function

Point spread function (PSF) analysis shows where the point sources of the optical system are shifted. The contrast detection limit within a PSF is determined by the photon noise and speckle noise in the image [16]. Figure 4 shows the PSF analysis result of the lens system. According to the analysis result of the obtained design, it is revealed that the lens is focused with minimal propagation of point shifts and provides high resolution. This is shown by the fact that the relative irradiance value reaches 1 in the curve.

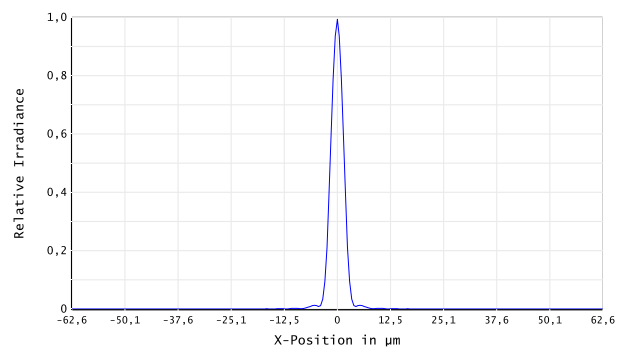

**Figure 4**. Point Spread Function

As a result, Figure 4 shows the response of an optical system to a point source and shows how a single point of light spreads across the image plane.

**4.3. Spot Diagram**

The spot diagram shows the focusing ability of the optical system and the aberrations in the image. Figure 5 shows the spot diagrams of the lens system. It has a high focusing ability in the 0.00 (deg), 0.60 (deg) and 1.20 (deg) areas. It indicates the angular areas that the optical system can image. The IMA value indicates the distance of the imaging area at these angles to the point formed on the image plane of the optical system. The results obtained show that the lens successfully focuses the rays coming from different angles to the points specified in the design phase and provides high accuracy with low deviation values. In addition, the small RMS and GEO radius values obtained in the analysis and shown in the graph mean that the optical system works with very good focusing and low aberration. As a result, the RMS radius of the three fields of view is quite smaller than the airy disk radius. This shows that the effect of improving the aberrations is significant.

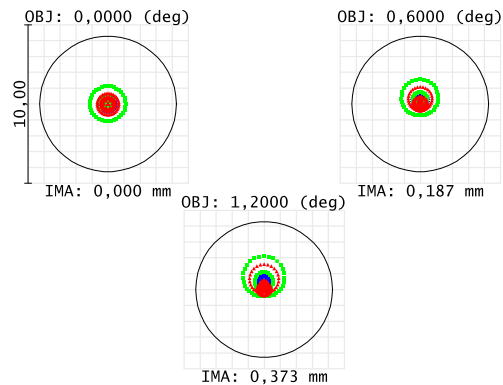

**Figure 5.** Spot Diagram.

As a result, Figure 5 represents the distribution of light from a point source as it passes through an optical system and shows the size and location of diffraction-limited spots at various points along the image plane.

**4.4. Seidel Diagram**

Seidel diagram is a diagram that shows the optical aberrations in the designed lenses according to their individual surfaces. According to the diagram obtained in Figure 6, optical aberrations are shown according to individual surfaces. As a result of the optimizations made in the SUM section at the end of the diagram, the optical aberrations transferred from the first surfaces were minimized.

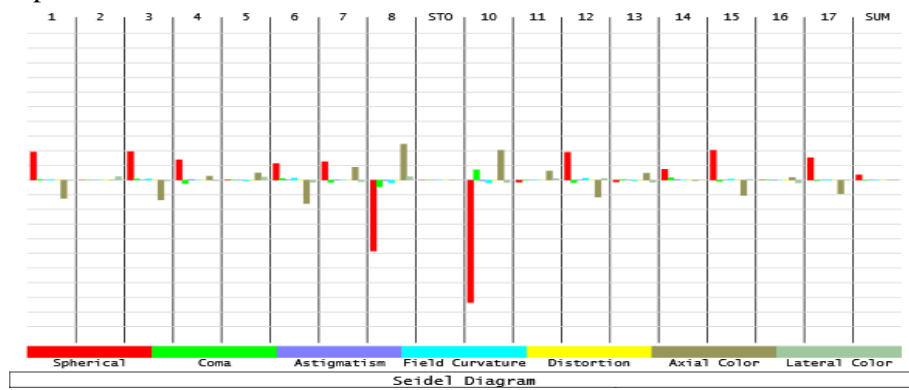

**Figure 6.** Seidel diagram.

The Seidel diagram in Figure 6 visualizes the primary aberrations (spherical, coma, astigmatism, field curvature and distortion) of an optical system.

**4.5. Field Curvature and Distortion**

The analysis of field curvature and distortion, presented in Fig. 7, demonstrates the deformation within the design. Significant field curvature and distortion can lead to poor image quality and positional errors for moving targets, while also negatively affecting the seamless mosaicking of component images [17]. The design analyses following the optimizations demonstrate that the relay lens delivers high performance in military optical systems. The maximum distortion was found to be 0.0118%, a notably low value that indicates the design achieves exceptionally high resolution with remarkably low distortion.

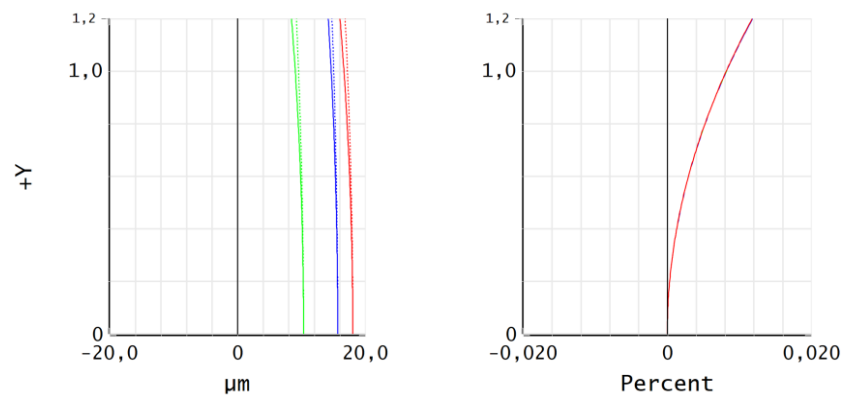

**Figure 7.** Field Curvature and Distortion.

Field curvature describes the change in focus along the image plane where a flat object appears curved in the image. Distortion refers to the change in shape of an image, often causing straight lines to appear curved.

The Seidel diagram analysis indicates that aberrations in the optical system have been minimized. The PSF cross-sectional image demonstrates the system's high focal point accuracy. The MTF analysis reveals that the system performs well over a wide frequency range, achieving significant modulation transfer values. Furthermore, the spot diagram confirms the optical system's focusing precision and image quality.

**4.6. Image Analyses**

These analyses are commonly used to identify, measure, and interpret objects, features, patterns, and structures within an image. Fig. 8 presents the extended diffraction image analysis graph, which illustrates how the optical system processes an image through diffraction. In the image, a large "F" character is represented by varying intensity levels, reflecting the frequency and sharpness of the lens system. The details of the character are clearly visible. Diffraction-induced spreading may cause slight blurring at the edges of the image. However, the edges of the character remain clearly visible, which indicates that the diffraction effect is minimal and the details are well-preserved. The color scale on the graph represents the normalized values of light intensity and contrast (0.0–1.0). Along the edges of the character, the intensity ranges between 0.6 and 1.0, indicating high contrast transfer. This demonstrates that the system provides clear imaging, particularly at low and medium spatial frequencies. The simulation was scaled to produce an image with a resolution of 4096 × 4096 pixels, enabling a high-resolution assessment. Diffraction-induced spreading may cause slight blurring at the edges of the image. However, the edges of the character remain clearly visible, indicating that the impact of diffraction is reduced. In the geometric image analysis presented in Fig. 9, it is shown how the optical system is geometrically imaged and that the system is not affected by geometric distortions. The efficiency value of 100.00% observed in the figure indicates that the system utilizes energy very efficiently, with all energy contributing to the image formation process.

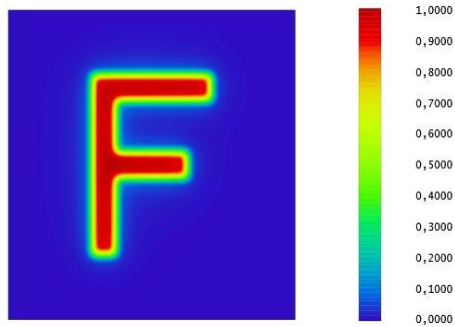

**Figure 8.** Diffraction Image Analysis

Figure 8 depicts the diffraction-limited performance of the lens system, highlighting its ability to produce high-quality and sharp images.

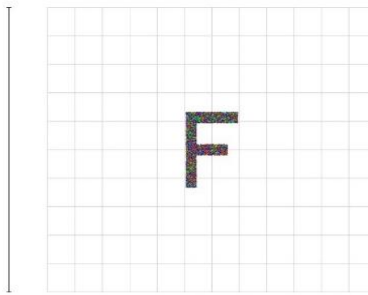

**Figure 9.** Geometric Image Analysis.

This analysis examines how well the system preserves the geometry of objects, including measurement of distortions, field curvature, and other geometric deviations.

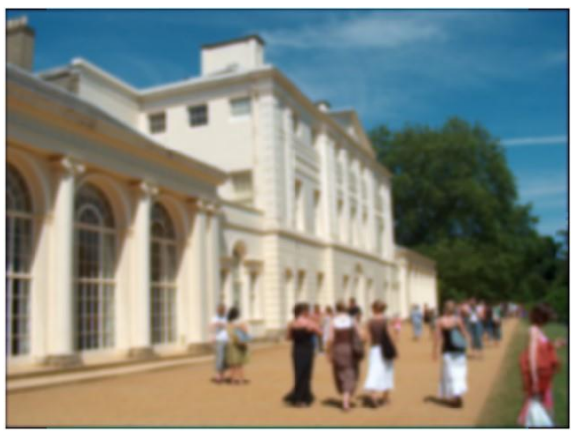 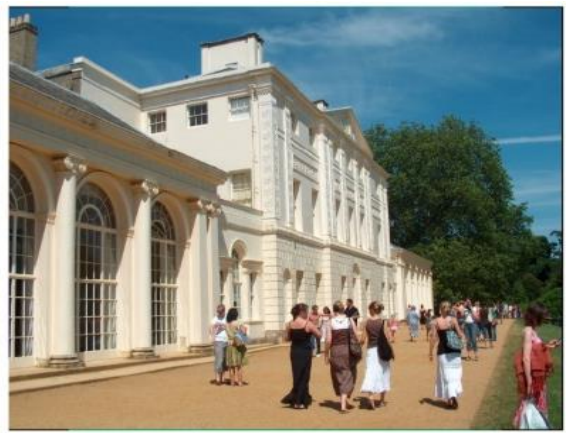

a) Before Optimization          b) After Optimization

**Figure 9.** Demo Image Analysis Before and After Optimization

Demo Image Analysis demonstrates the performance of the optical system through real-world imaging. The demo image analysis shown in Figure 10 is a basic reference for evaluating the performance of the design. By analyzing the clarity of details, sharpness, distortion and deviations in the image, it is understood whether the design meets the targeted criteria.

**4.7. Merit Function**

The merit function is a numerical parameter used to evaluate the conformity of an optical system to specific performance criteria. It serves as an error metric, indicating how closely the design aligns with the desired optical performance. Typically, it is employed to minimize deviations from target values. As the merit function value approaches zero, the design becomes more ideal and better aligned with the specified parameters.

The merit function value for our design is shown to be 0.0144008. This value demonstrates that the design is very close to meeting the targeted performance criteria and that design errors have been minimized. A low merit function value indicates that the optical system possesses high resolution and achieves the desired optical quality. This result highlights the design's success in terms of optical performance and confirms that it has been optimized to meet the targeted imaging specifications.

Merit function combines various criteria such as deviations and image quality into a single value that guides the optimization process to achieve the best possible design.

**4.8. Optical Path Difference**

The Optical Path Difference (OPD) Fan plot is used to measure and quantify deviations of light waves from the ideal wavefront in the optical system. In an ideal scenario, a wavefront is a surface where light waves are in phase, meaning their optical paths are identical. In practice, however, this ideality is unattainable due to inherent material properties and design limitations, which result in deviations from the ideal wavefront. Fig. 12 presents the OPD Fan plot for our optical system. The analysis shows that the OPD deviations are confined within ±0.1 waves, highlighting the system's superior optical performance with minimal deviations. For an aberration-free system, the aberration curve should be a straight line coinciding with the x-axis [18]. The OPD Fan plot of our design indicates that the $P_x$ and $P_y$ curves exhibit a generally flat and consistent profile, demonstrating that the aberrations and field curvature are well corrected and maintained at minimal levels. Moreover, the consistent OPD profiles across varying field angles (0°, 0.6°, 1.2°) underscore the system's ability to deliver robust performance over a wide field of view.

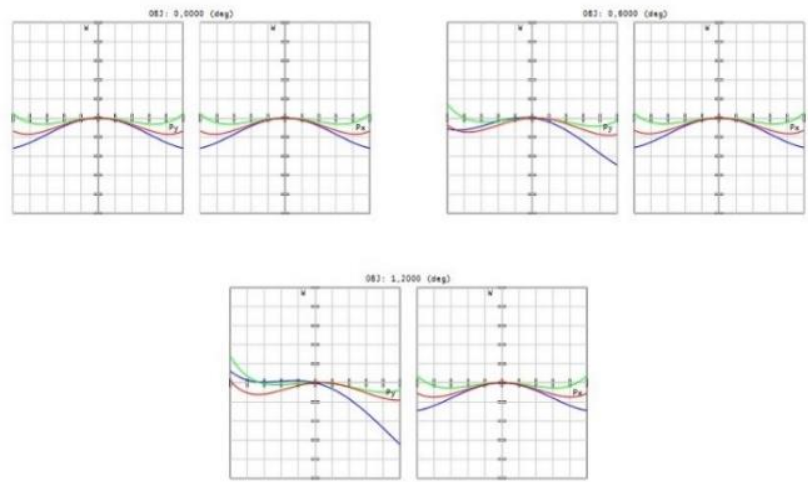

**Figure 12.** Optical Path Difference Fan Analysis.

Figure 12 evaluates the performance of the system in terms of deviations and determining the changes in optical path lengths and phase differences throughout the optical system.

## 5. Results and Discussions

As a result, the analyses performed show that the designed optical system meets the targeted performance criteria and the design is successfully optimized. According to the MTF analysis, sufficient contrast transfer is provided, and the PSF analysis confirms that the lens focuses the point shifts with minimal spread, achieving high resolution. In addition, the spot diagram shows that the lens focuses the rays coming from different angles to the points specified in the design phase with low deviation values and provides high accuracy. The Seidel diagram shows that the optical defects on the lens surfaces are minimized.

The maximum distortion value remains at a low level of 0.0118%, confirming that the design achieves high resolution and low distortion. The Extended Diffraction Image Analysis results show that the system preserves high-resolution details and that the diffraction effect is at a minimum level. The geometric image analysis reveals that the optical system operates with high efficiency and without geometric distortions. The effective focal length (EFFL) value of the system in the design is approximately 9.5 mm. In optical systems, the EFFL value directly affects the imaging of the system and the distance between the target and the sight. The effective focal length of 9.5 mm is especially suitable for compact and lightweight optical systems. In portable systems such as rifle sights, this focal length provides a wide personal field of view. This wide field of view offers advantages in terms of rapid target acquisition and aiming. Another parameter in the design is the total optical path (TOTR). The TOTR value of our system is approximately 20 mm. The TOTR value determines the overall dimensions and portability of the optical system. The total length of 20 mm maintains the compactness of this design. Especially in military applications, the lightness and ease of use of the sight are essential. This design meets these features. The aperture value allows the system to collect light and, therefore, affects the image brightness. While the aperture is 3 mm in size, it provides sufficient light collection ability, and at the same time, the system has a positive effect on the depth of field. This provides sharp focus on the target, especially at different distances, which is a critical feature for rifle sights. The high values obtained in the MTF analysis reveal that the design can provide the contrast required for sighting applications. This helps to accurately detect the target by providing clear and sharp images. The low merit function value confirms that the optical system has high resolution and provides the desired optical quality. All these findings emphasize that the designed optical system can be used as a high-performance relay lens for applications such as target acquisition and tracking in military optical systems. According to the optical path difference analysis, the system showed good optical performance with minimal aberrations. The consistent profiles of the OPD values showed that the system has robust performance across a wide field of view.

## 6. Recommendations

These design parameters and analysis results demonstrate that the system is suitable for weapon-mounted sights requiring precise aiming at short distances. For applications involving longer distances, the focal length would need to be increased. To enhance performance under low-light conditions, the aperture size should be optimized accordingly. While implementing these solutions, the results of the MTF analysis must confirm the adequacy of the design in terms of contrast and resolution. This requirement should also be validated by other analyses.